\documentclass{article}
\usepackage{spconf,amsmath,graphicx}
\usepackage{booktabs}

\title{Robust Wind Turbine Blade Segmentation from RGB Images in the Wild}

\name{Raül~Pérez-Gonzalo$^{1,2}$, Andreas~Espersen$^{2}$, Antonio~Agudo$^{1}$\thanks{This work has been supported by the Innovation Fund Denmark under 2021 ID1044-0044A and by the project MoHuCo PID2020-120049RB-I00 funded by MCIN/AEI/10.13039/501100011033.}}
\address{$^{1}$Institut de Robòtica i Informàtica Industrial, CSIC-UPC, Spain\\$^{2}$Wind Power LAB, Copenhagen, Denmark}
\begin{document}
%
\maketitle
\begin{abstract}
With the relentless growth of the wind industry, there is an imperious need to design automatic data-driven solutions for wind turbine maintenance. As structural health monitoring mainly relies on visual inspections, the first stage in any automatic solution is to identify the blade region on the image. Thus, we propose a novel segmentation algorithm that strengthens the U-Net results by a tailored loss, which pools the focal loss with a contiguity regularization term. To attain top performing results, a set of additional steps are proposed to ensure a reliable, generic, robust and efficient algorithm. First, we leverage our prior knowledge on the images by filling the holes enclosed by temporarily-classified blade pixels and by the image boundaries. Subsequently, the mislead classified pixels are successfully amended by training an on-the-fly random forest. Our algorithm demonstrates its effectiveness reaching a non-trivial 97.39\% of accuracy. 
\end{abstract}

\vspace{-0.15cm}
\section{Introduction} 
\vspace{-0.15cm}

Wind energy has demonstrated to be an excellent alternative energy source, thanks to being completely renewable and environmentally sustainable~\cite{windreport}. The increase in installed capacity urges to adapt to more demanding challenges in terms of cost and logistics~\cite{30capacity}, with special focus on developing non-destructive inspection manners for health monitoring. Drone inspections have become a fast and reliable setup for wind turbine diagnosis~\cite{drone}, capturing around 400 high-resolution images per turbine to design a repair plan. All in all results in high-volume data that encourages the development of automated, scalable and cost-effective solutions. 

To successfully design a data-driven maintenance framework for wind turbines in operation, current researchers are tackling distinct image-based problems such as blade defect detection~\cite{aidrone,maskrcnn} or autonomous navigation~\cite{realtime, automateddrone}. However, current solutions have several limitations and are not being applied yet to the industry. Hence, these challenges must be simplified by means of distinct low-level vision techniques. To this aim, this paper proposes addressing segmentation to reduce the complexity of these learning tasks.

\begin{figure}[t!]
  \centering
\resizebox{8.4 cm}{!} {
  \centerline{\includegraphics[width=8.5cm]{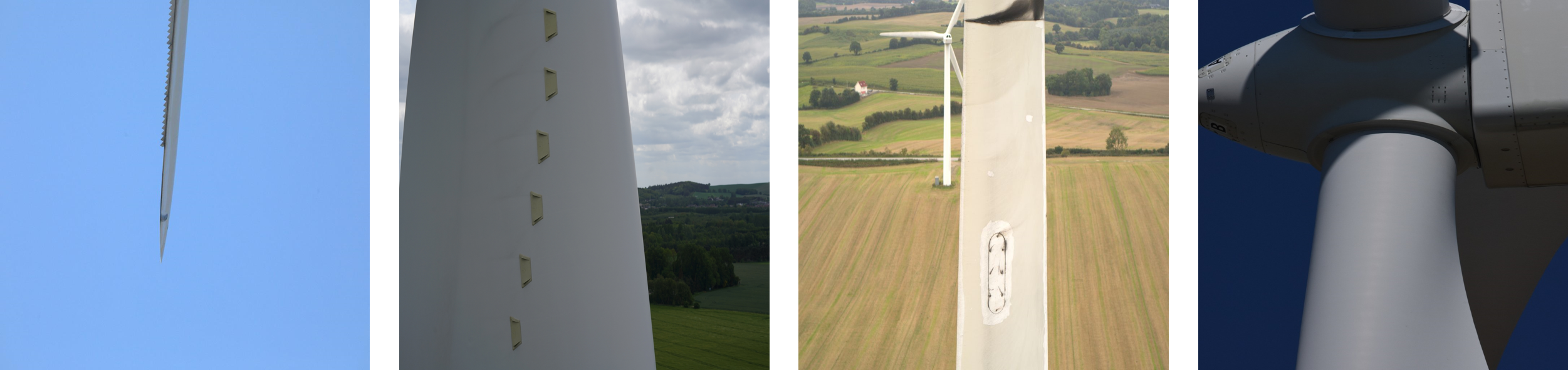}}}
  \vspace{-0.3cm}
\caption{\textbf{Wind turbine blade images in the wild} with distinct complexity due to the variation of the landscapes, the blade shape and size, and the presence of shadows or dirt.}
\label{fig:introd}
\vspace{-0.55cm}
\end{figure}

Current state-of-the-art segmentation methods rely on learning-based approaches, usually employing an encoder-decoder architecture~\cite{deeplab,sw,cvpr2022} and, recently, an attention mechanism on the encoder~\cite{bmvc,attention2}. With a properly tailored architecture, these models have proven to successfully obtain high precision results. Unfortunately, they do not provide a segmentation algorithm that generalizes for newly acquired images~\cite{generic}. A generic solution is essential to deal with a large variety of images (see Fig.~\ref{fig:introd}) obtained on regular wind turbine inspections under a different setup. 

A common technique to enhance the segmentation mapping is including a post-processing step. DeepLab~\cite{deeplab} utilized Conditional Random Fields to fine-tune the localized segmented objects. DeepLab V3+~\cite{deeplabv3+} is further strengthened through superpixel merging in~\cite{generic}. Another interesting approach is ResNeSt~\cite{resnest}, which last stage incorporates a channel-wise attention with multi-path representation. However, these methods do not leverage our prior information about blade images, such as that the blade traverses the image or that images of the same inspection share image properties.  

This work proposes a novel segmentation algorithm that identifies the blade region for very diverse images in appearance. First, a preliminary mask is obtained by a customized U-Net model~\cite{unet}, whose architecture still inspires current techniques~\cite{unetformer}. Then, this preliminary result is improved through a set of post-processing steps: an extended hole filling routine from~\cite{fill} followed by an unsupervised random forest~\cite{randomforest}; refined again through another hole filling step. The hole filling relies on properly identifying the blade image borders and the random forest on underfitting the preliminary masks from the same blade surface. As a result, we obtain an accurate, robust and efficient model for blade segmentation that can handle a wide variety of images in the wild.

\section{Wind Turbine Blade Segmentation}
\label{section_methods}

Let $\mathbf{I}$ be a $H \times W \times 3$ color image of a wind turbine blade that was captured from an arbitrary point of view and under uncontrolled lighting conditions. Our goal is to recover a $H \times W$ binary mask $\hat{\mathbf{S}}$ to indicate the region in the image that belongs to the blade. In our formulation, the mask is represented by the binary entries $\hat s_{h,w}$, where $h=\{1,\ldots,H\}$ and $w=\{1,\ldots,W\}$ denote the pixel coordinates in the mask.

Given the corresponding ground-truth mask $\mathbf{S}$, we propose a supervised learning framework composed of four modules to learn the mapping $\mathbf{I}_{i,j}\rightarrow{\mathbf{S}_{i,j}}$ for a $i$-th blade and $j$-th  image: 1) an encoder-decoder network algorithm with a tailored loss, 2) a hole filling step, 3) a random forest block and, 4) a latter hole filling step to refine the solution. Without loss of generality, for every blade surface there are considered up to $J$ images for the $i$-th blade.


\subsection{Blade U-Net (BU-Net)}

Given an input RGB image $\mathbf{I}$, our BU-Net aims at classifying each input pixel into blade or background, i.e., obtaining the output mask ${\hat{\mathbf{S}}}^{BU}$ by exploiting the labels in $\mathbf{S}$. This module follows a standard U-Net architecture~\cite{unet}, and its last layer includes a $1 \times 1$ convolution to map the feature channels to the desired number of classes. The mask ${\hat{\mathbf{S}}}^{BU}$ is obtained by combining four predictions via soft voting, where each one corresponds to a flipping rotation. Additionally, a regularized focal loss is employed to adapt the learning task to our particular dataset. The two terms are detailed below.

The \textit{Focal Loss} \label{sec:focal-loss} optimizes our segmentation mapping~\cite{focal} by penalizing low-confidence predictions. Moreover, to avoid benefiting the majority class (in our case, the background) relative to the minority class, we apply weights to each class: 

\vspace{-0.2cm}
\begin{gather} 
\label{focal_loss}
    \mathcal L_{f}(\mathbf{S}, {\hat{\mathbf{S}}}^{BU}) = -\sum_{h,w} \left[ \alpha \left(1-\sigma(\hat s^{BU}_{h,w})\right)^\gamma s_{h,w}\log\sigma(\hat s^{BU}_{h,w})  \right. \nonumber \\ 
   \left.  + (1 - \alpha) \left(\sigma(\hat s^{BU}_{h,w}))\right)^\gamma (1-s_{h,w})\log(1-\sigma(\hat s^{BU}_{h,w}))    \right],
\end{gather}
where $\sigma(\cdot)$ denotes the sigmoid function, $\gamma$ controls the confidence penalty; and $\alpha$ sets the class balance.  




The \textit{Contiguity Loss}~\cite{cont} minimizes the jaggies in the blade boundaries, forcing to detect contiguous objects:
\begin{align}
    \mathcal L_{c}({\hat{\mathbf{S}}}^{BU}) = \frac{1}{HW} \Big[ \sum_{h,w} \left(   \sigma(\hat s^{BU}_{h+1,w})-\sigma(\hat s^{BU}_{h,w})  \right)^2 \nonumber \\
+ \sum_{h,w} \left(    \sigma(\hat s^{BU}_{h,w+1})-\sigma(\hat s^{BU}_{h,w})   \right)^2 \Big]^{1/2} .
\end{align}

The \textit{Total Loss} is defined as a linear combination of the two previous losses modulated by the scalar $\lambda$:
\begin{equation}
\label{total_loss}
\mathcal L = \mathcal L_{f} + \lambda\mathcal L_{c}\text{ } .
\end{equation}
 As the output of this block is the probability for the blade class, i.e., $\sigma(\hat{\mathbf{S}}^{BU})$, it is quantized by a threshold of 0.4.

\subsection{Hole Filling}
\label{hole_filling}


The estimated segmentation $\hat{\mathbf{S}}^{BU}$ still contains some non-realistic regions, including some holes. We propose a {\em hole filling} step to reduce these artifacts. Our approach is able to not only fill in the holes surrounded by blade pixels, but also when they are surrounded by the image borders. By exploiting a spatial prior, our approach relies on localizing first the blade pixels in the image boundaries, i.e., the blade-borders.

\begin{figure}[t!] 
\begin{minipage}[b]{1.0\linewidth}
  \centering
  \resizebox{8.5 cm}{!} {
  \centerline{\includegraphics[width=8.5cm]{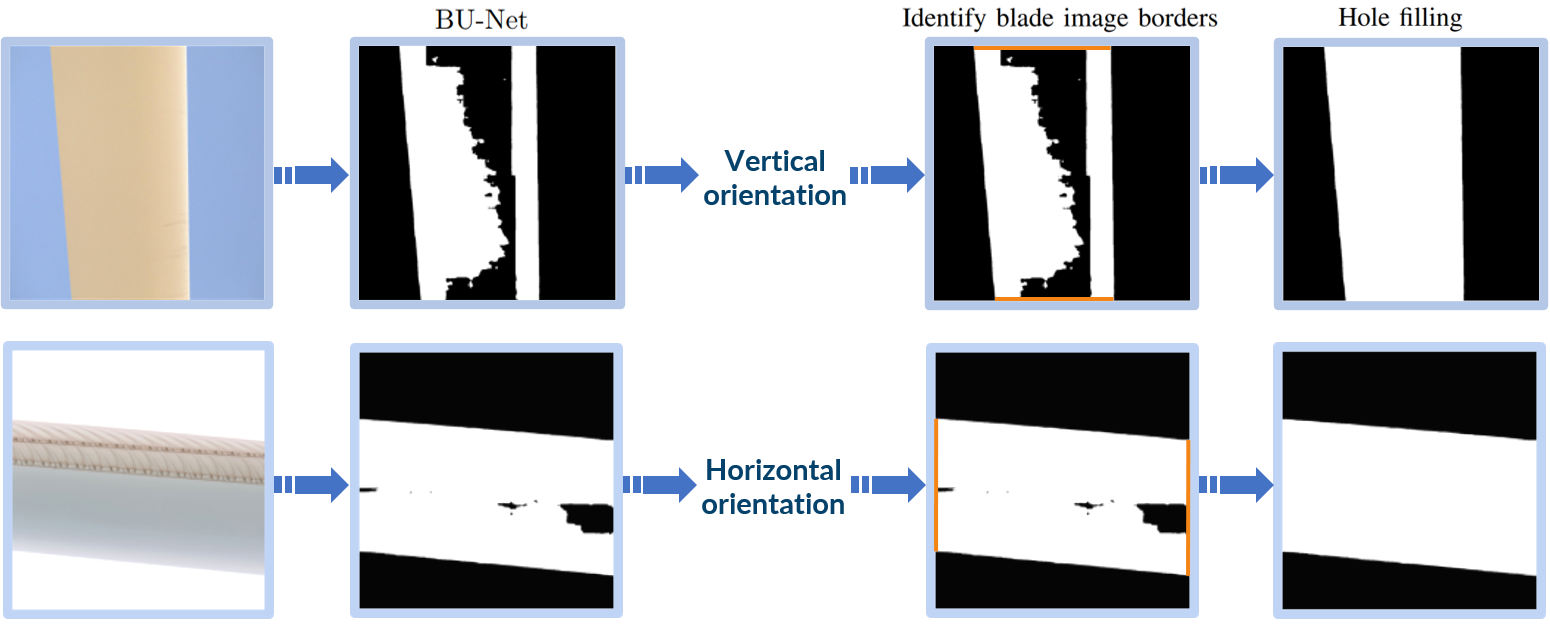}}}
  \vspace{-0.25cm}
\end{minipage}
\caption{\textbf{Hole filling representation.} First, the blade orientation is detected to identify the blade image borders (in orange). Then, holes that edge with the image borders are filled. }
\label{fig:holefilling}
\vspace{-0.3cm}
\end{figure}

Assuming that the blade traverses the image, we first compute the blade orientation by means of an accumulated gradient. Basically, it is classified as vertical (horizontal) if the accumulated gradient along the $x$-axis is higher (lower) than along the $y$-axis. Once the orientation is identified, we know the blade-border locations, so we can enforce their continuity by setting their pixels as blade. Finally, we proceed with the standard hole filling algorithm~\cite{fill} to obtain the output $\hat{\mathbf{S}}^{H}$. Both vertical and horizontal orientation examples are displayed in Fig.~\ref{fig:holefilling}. The hole filling step is applied twice: 1) considering as input the estimation in our BU-Net and, 2) using as input the output of our random forest algorithm that is introduced below. To differ between both hole filling estimations, we will use the notation $\hat{\mathbf{S}}^{H1}$ and $\hat{\mathbf{S}}^{H2}$, respectively.


\vspace{-0.2cm}
\subsection{Unsupervised Random Forest}
\label{random_forest}

Finally, we propose a simple random forest to denoise the estimation and to fix uncommon errors of a single image. The inability of previous steps to capture global context is tackled by learning which RGB pixels are usually mapped as foreground, providing a fine-grained boundary precision. To this end, we learn the mapping $\{\mathbf{I}_{i,j}\}_{j \in J} \rightarrow \{\hat{\mathbf{S}}^{H1}_{i,j}\}_{j \in J}$ given the $i$-th blade surface. This model ensembles tree estimators by averaging their probabilistic predictions to set whether a single RGB pixel and its local neighborhood is foreground:

\vspace{-0.5cm}
\begin{equation}
\label{eq:rf} 
P^{RF}\big(\hat s_{h,w}^{RF} \big) = \frac{1}{|RF|} \sum_{t \in RF} P^t\big( \hat s_{h,w}^{t} | \hat s_{h,w}^{H1}, \text{neighbors}(\hat s_{h,w}^{H1}) \big)\text{ } , \nonumber
\vspace{-0.2cm}
\end{equation}
where $RF$ is the set of decision trees $t$. It is worth pointing out that this approach enables our algorithm to adapt to the image features of a specific $i$-th blade. Once the model is trained, it is applied to obtain the estimation $\hat{\mathbf{S}}^{RF}_{i,j}$.

Lastly, we again apply a hole filling (section~\ref{hole_filling}) step to refine the solution, obtaining $\hat{\mathbf{S}}^{H2}$ as our final estimation $\hat{\mathbf{S}}$.



    




\section{Experimental Results} \label{sec:results}


In this section, we demonstrate the efficacy of our segmentation approach by conducting comprehensive quantitative and qualitative analysis. Additionally, we present a comparative assessment against competing techniques, offering detailed results for each step implemented within our algorithm.


\vspace{-0.5cm}
\subsection{Dataset} \label{sec:data} 
\vspace{-0.1cm}

We present a dataset that consists of high-resolution images taken by drone and ground-based equipment from distinct locations of the blade. The images provide a large variety of blade sizes, shapes, illumination conditions, as well as the background to be observed (Fig.~\ref{fig:introd}). Another background variation arises from on-shore and off-shore wind turbines. For this challenging dataset, we propose 1,712 images for training, 120 for validation and 200 for test. As a segmentation ground-truth is not provided, we manually annotated all the images. To make a fair analysis, the three sets are composed of different windfarms and inspection campaigns, ensuring their independence and that the test properly generalizes. In particular, we randomly selected 20 images of each windfarm to generate the validation and test sets. On the other hand, the training data was selected from a pool of different blade images, prioritizing the ones that are more challenging. These are the images in the root, the tip or the max-cord (where the blade section is widest). All color images and masks were min-max normalized. In addition to that, flipping and cropping strategies were applied to learn the desired invariance and robustness properties.


\vspace{-0.5cm}
\subsection{Implementation Details} \label{sec:training-seg}
\vspace{-0.1cm}

The original input images were resized to $1024 \times 1024$ pixels. We use Adam solver~\cite{adam} with an initial learning rate of $10^{-4}$ and a mini-batch of 1. A custom scheduler is employed to reduce the learning rate when there is no improvement on the validation loss after three epochs. Early stopping is performed. The model is trained using a NVIDIA GeForce RTX 3080 GPU. The weights for the focal loss in Eq.~\eqref{focal_loss} are set to $\gamma=2$ and $\alpha=0.25$, and the weight in Eq.~\eqref{total_loss} $\lambda=1$.

\vspace{-0.5cm}
\subsection{Quantitative Evaluation}
\vspace{-0.1cm}



The segmentation performance per image is analyzed in terms of accuracy, recall and F1-score. 







\textbf{Hyperparameter Search.} Our random-forest step (section~\ref{random_forest}) is optimized in terms of the number of split branches and the number of decision trees, obtaining the highest accuracy with five tree estimators that have a maximum of four split branches. Additionally, when the local pixel neighbors are included in the input, we enhance the random forest masks. However, increasing the number of input neighbors does not increase the segmentation accuracy, thus, we took as input the local pixel and its most immediate neighbors. 

\textbf{Test Set Results.} The performance over the test set is compared between each algorithm step. In particular, we report the accuracy, recall and F1-score in Table~\ref{tab:seg-accuracy}. 

\vspace{-1.5cm}
By only employing our BU-Net, we obtain a notable outcome of 92.28\% of accuracy. However, this result is not sufficiently good for blade assessments, because the recall of 85.73\% could compromise the detection of defects. For instance, if a large area of the blade is masked as background, it could cover the region where there are structural damages, which is a severe defect that cannot be overlooked.

By incorporating the first hole-filling step, we substantially improve performance. Namely, the recall increases to 92.08\%, which represents the highest gain of all the post-processing steps. This huge difference is due to the BU-Net being able to successfully capture the blade borders, but not the inner region of the blade. Further algorithm steps are necessary to yield further improvement, reaching in the end a novel performance of 97.39\% of accuracy and 93.35\% of recall. On balance, and as it can be seen in Table~\ref{tab:seg-accuracy}, every step is key to sort out the problem properly, obtaining a final accurate solution that cannot be achieved directly by our BU-Net. 

\begin{figure}[t!] 
\begin{minipage}[b]{1.0\linewidth}
  \centering
  \resizebox{8 cm}{!} {
  \centerline{\includegraphics[width=8.5cm]{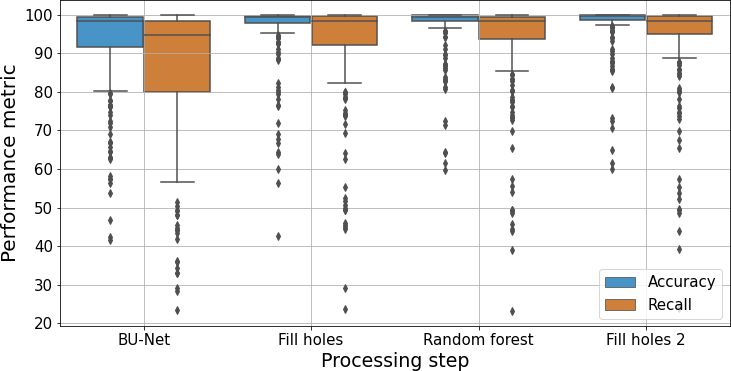}}}
  \vspace{-0.35cm}
\end{minipage}
\caption{\textbf{Box plot test results after each segmentation step.} }
\label{fig:boxplot}
\vspace{-0.4cm}
\end{figure}

Fig.~\ref{fig:boxplot} analyzes the performance dispersion of each step. We observe the skewness of the BU-Net performance: rather spread along the tail with lower performance, containing many instances with poorer accuracy and recall compared with their median. As we include the different steps in our algorithm, that dispersion is reduced notably. Therefore, despite not substantially improving the mean performance, the subsequent steps are the cornerstone to ensure robustness.

\begin{table}[t!]
\centering
\caption{\textbf{Test results after each segmentation step.} }
    \vspace{-0cm}
    \label{tab:seg-accuracy}
\resizebox{8.5 cm}{!} {
\begin{tabular}{ccccc}
\toprule
     \multicolumn{1}{c}{Performance} & \multicolumn{1}{c}{BU-Net} &   \multicolumn{1}{c}{Fill holes}  &  \multicolumn{1}{c}{Random forest}  &  \multicolumn{1}{c}{Fill holes 2}\\
    \multicolumn{1}{c}{metric} &  \multicolumn{1}{c}{${\hat{\mathbf{S}}}^{BU}$} &   \multicolumn{1}{c}{$\hat{\mathbf{S}}^{H1}$} &  \multicolumn{1}{c}{$\hat{\mathbf{S}}^{RF}$} &  \multicolumn{1}{c}{$\hat{\mathbf{S}}^{H2}$} \\ \midrule
   Accuracy (\%) &  92.28 & 96.11 & 97.11 & 97.39\\
   Recall (\%) &  85.73 & 92.08 & 92.37 & 93.35 \\
   F1-score (\%) & 90.41 & 94.53 & 95.12 & 95.73 \\
   mIoU (\%) & 87.71 & 93.57 & 95.07 &  95.53 \\
\bottomrule
\end{tabular}}
\vspace{-0.25cm}
\end{table}

\begin{table}[t!]
\centering
\caption{\textbf{Quantitative comparison on blade segmentation.} Relative metrics are with respect to the worst estimation.} 
\vspace{0cm}
\label{tab:seg-unet-compare}
\resizebox{8.5 cm}{!} {
\begin{tabular}{ccccccc}
\toprule
     \multicolumn{1}{c}{Method} & \multicolumn{1}{c}{Accuracy}  & \multicolumn{1}{c}{Recall} & \multicolumn{1}{c}{F1-score} & \multicolumn{1}{c}{mIoU} & Relative & Relative\\
    \multicolumn{1}{c}{} & {\%} & {\%}  & {\%}  & {\%} & accuracy & recall \\
    \midrule
   U-Net~\cite{unet} & 86.24 & 68.93 & 77.95 & 79.94 & 1 & 1\\ 
   DeepLabv3+~\cite{deeplabv3+}  & 94.14 & 87.38 & 89.03 & 90.74 & 1.09 & 1.27 \\ 
   SW~\cite{sw} & 93.48 & 91.71 & 91.37 & 89.71 & 1.08 & 1.33 \\ 
   ResNeSt~\cite{resnest} & 94.23 & 91.47 & 92.77 & 91.20 & 1.09 & 1.33 \\ 
   U-NetFormer~\cite{unetformer} & 96.20 & \textbf{93.51} & 94.42 & 93.65 & 1.12 & \textbf{1.36} \\ 
   Ours $\hat{\mathbf{S}}$ & \textbf{97.39} & 93.35 & \textbf{95.73} & \textbf{95.53} & \textbf{1.13} & 1.35 \\ 
\bottomrule
\end{tabular}}
\vspace{-0.5cm}
\end{table}

\textbf{Comparative Results.} As the BU-Net employs the U-Net architecture~\cite{unet}, we compared our proposed algorithm with the original one and state-of-the-art methods (Table~\ref{tab:seg-unet-compare}). We outperform those methods, thanks to overcoming the blade-background imbalances, and escaping from the tendency of predicting the most common region. 

\vspace{-0.3cm}
\subsection{Qualitative Results}
\vspace{-0.1cm}

\begin{figure}[t!]
\resizebox{8.5 cm}{!} {
\hspace{-0.1cm}\begin{tabular}{@{}c|c@{}}
\includegraphics[width=0.4\textwidth]{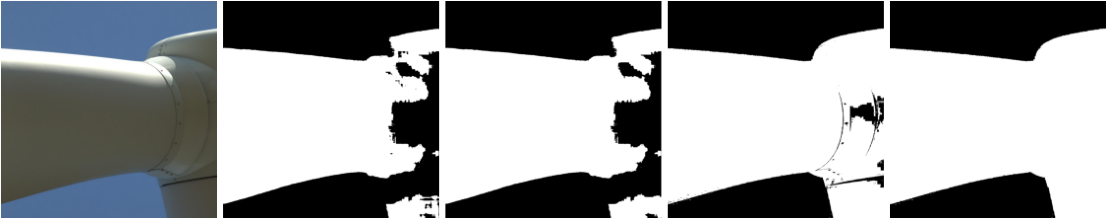}&
\includegraphics[width=0.4\textwidth]{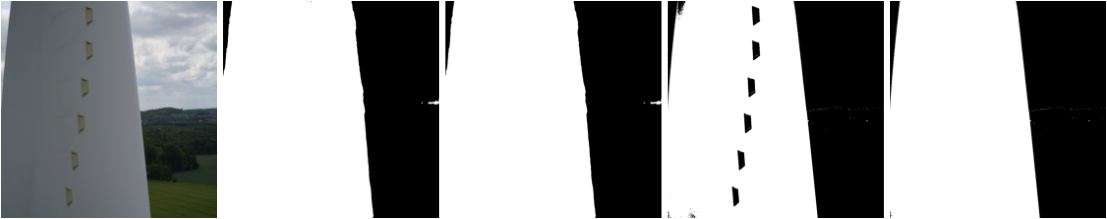}\\
\includegraphics[width=0.4\textwidth]{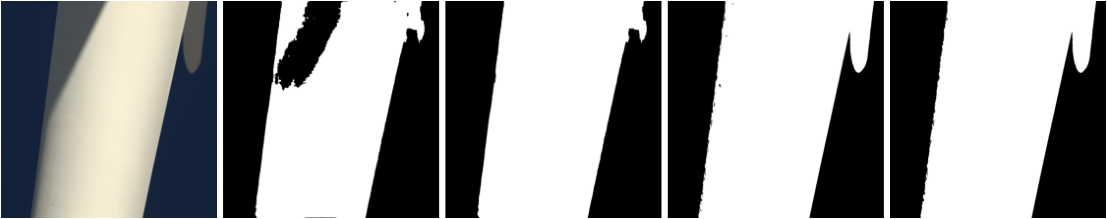}&
\includegraphics[width=0.4\textwidth]{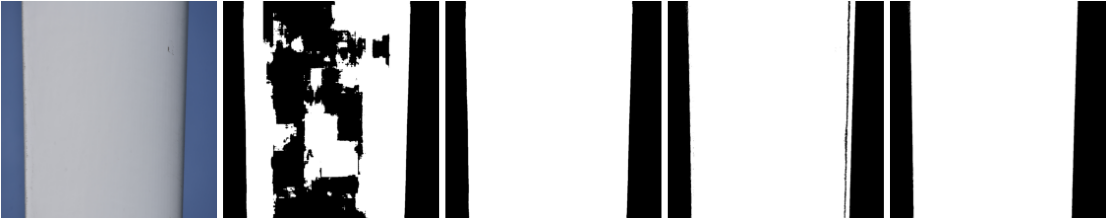}\\
\includegraphics[width=0.4\textwidth]{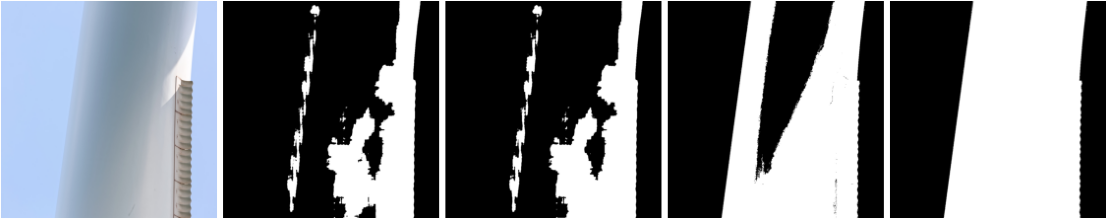}&
\includegraphics[width=0.4\textwidth]{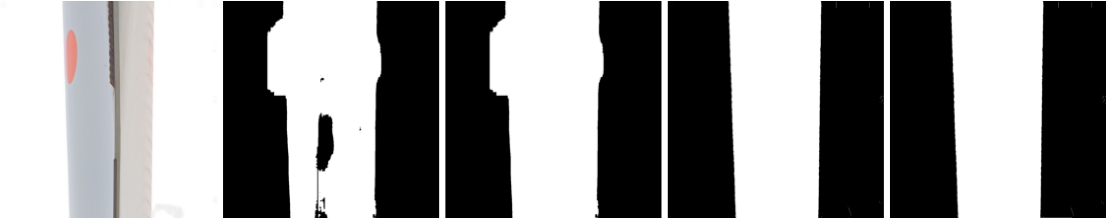}\\
\end{tabular}}
\vspace{-0.2cm}
\caption{\textbf{Qualitative evaluation on test images after applying each step of our proposed segmentation algorithm.} On both sides, the same information is displayed. \textbf{First column:} Input color image ($\mathbf{I}$). \textbf{From second to fifth column:} BU-Net (${\hat{\mathbf{S}}}^{BU}$), first hole filling ($\hat{\mathbf{S}}^{H1}$), random forest ($\hat{\mathbf{S}}^{RF}$) and second hole filling ($\hat{\mathbf{S}}^{H2}\equiv\hat{\mathbf{S}}$) estimations, respectively.}
\label{fig:seg-visual}
\vspace{-0.35cm}
\end{figure}

We visualize the segmentation results of each algorithm step in Fig.~\ref{fig:seg-visual}. In general, our BU-Net can capture the blade edges and, therefore, its overall shape. However, it struggles with the inner area of the blade, which is addressed in the first hole filling step. Note that just using the standard hole filling approach would not improve the BU-Net outputs. Then, the random forest is responsible for fixing the misplaced pixels, acting as a regularizer of the pixel color intensity. By training a simple model, we ensure there is no overfitting and that it just learns the main patterns. Hence, if the BU-Net can in general classify which color intensities belong to the blade region, then the random forest would learn those patterns and resolve the particular images or pixels that failed, obtaining robustness in the solution. The second hole filling step is applied to solve the shortcomings the random forest can introduce. They are generally caused by dark regions that highly differ with the typical white color of the blade --a typical example is illustrated in Fig.~\ref{fig:seg-visual}, first row/right side--, where the random forest misplaces the vortex generators as background.


\begin{figure}[t!]
\resizebox{8.5 cm}{!} {
\hspace{-0.1cm}
\begin{tabular}{@{}c|c@{}}
\includegraphics[width=0.4\textwidth]{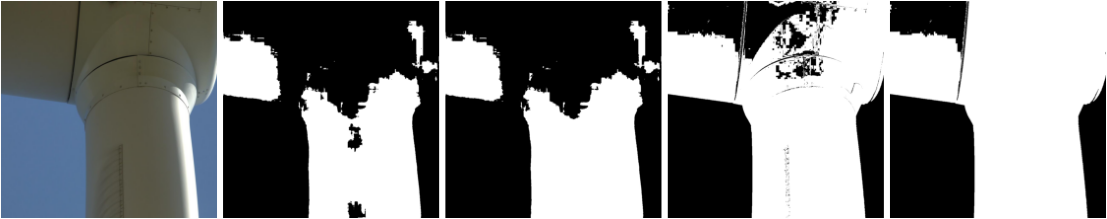}&
\includegraphics[width=0.4\textwidth]{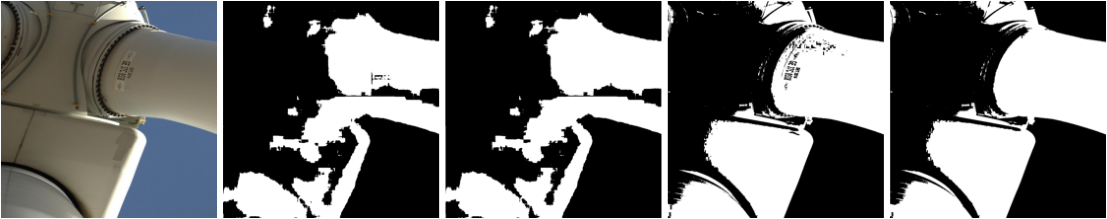}\\
\includegraphics[width=0.4\textwidth]{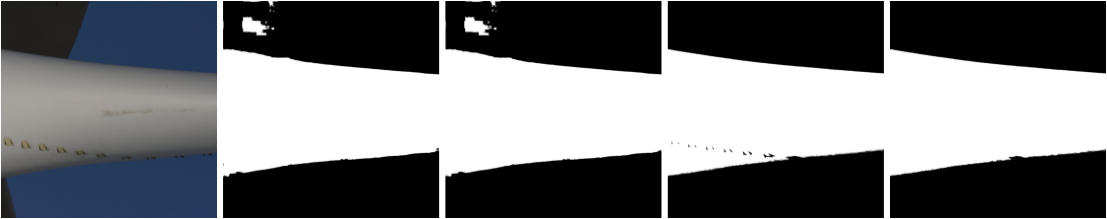}&
\includegraphics[width=0.4\textwidth]{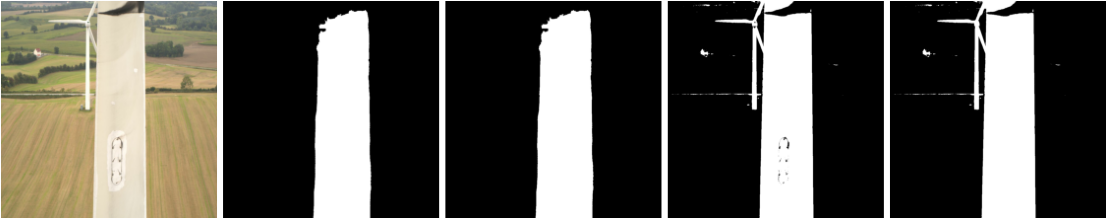}
\end{tabular}}
\vspace{-0.1cm}
\caption{\textbf{Failure cases on test images after applying each step of our segmentation algorithm.} See caption of Fig.~\ref{fig:seg-visual}.}
\label{fig:seg-fail}
\vspace{-0.0cm}
\end{figure}

In Fig.~\ref{fig:seg-fail}, we depict the most common cases where our segmentation algorithm could fail. To start with, the BU-Net might occasionally struggle with images of the wind turbine hub, because our dataset has a few instances of the hub. This is the case for the images in the first row. Distinct sources of contamination can also cause the BU-Net to mislead part of the blade with the background. This is the case for the image in the second row and right side. As the blade image borders are not fully captured and these images may contain a broad range of color intensities in the blade region, the hole filling and random forest steps are not sufficient to recover the blade region that has not been identified by the BU-Net.

In addition, an image could include wind turbines at its background or its own tower (see images in the second row). Ideally, we would like to predict the tower as blade, since it belongs to the turbine itself, but not the turbines in the landscape. This challenging problem is part of our future work.


\vspace{-0.2cm}
\subsection{Windfarm Dissimilarity}
\vspace{-0.1cm}



\begin{figure}[t!]
\centering
\includegraphics[width=3.1in]{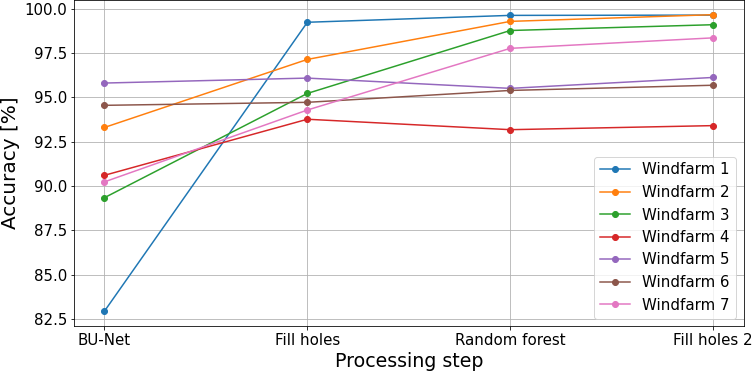}
    \vspace{-0.3cm}
    \caption{\textbf{Accuracy results over the test set for each windfarm.} It is included in each step in our method. For readability, the plot does not include every windfarm of the test set.}
    \label{fig:seg-inspections}
    \vspace{-0.3cm}
\end{figure}

Fig.~\ref{fig:seg-inspections} proves the generability of our approach, demonstrating that images from all the distinct windfarms have a high performance. Notice that this is not the case for the BU-Net without the post-processing steps, as it has really low performance for Windfarm \#1. Hence, it is the post-processing steps as a whole which ensures the generability. For instance, hole filling is crucial for Windfarm \#1, while random forest is crucial for Windfarm \#2 and \#3.
\vspace{-0.2cm}

\vspace{-0.1cm}
\section{Conclusion}
\vspace{-0.1cm}

In this paper we have presented a novel segmentation algorithm for wind turbine blade images. To this end, we have proposed a BU-Net model in combination with additional steps based on hole filling and random forest strategies that can solve the problem in an accurate and efficient manner. Our BU-Net exploits a focal loss that is regularized with a contiguity term. The hole filling is simple but effective, and helps to improve the solution after applying every learning model. Finally, our unsupervised random forest is trained on-the-fly capturing the main patterns, like a denoising step. Remarkable quantitative and qualitative results are provided on newly acquired blade images, even when they belong to distinct inspection campaigns and wind farms, validating the capability of our model to generalize properly. Our future work is oriented to extend our model to semantic segmentation where different parts of the wind turbine need to be segmented in the image. Furthermore, we would like to study its impact on facilitating the defect detection task.

\vfill\pagebreak

\bibliographystyle{IEEEbib}
\bibliography{References}

\end{document}